\documentclass[letterpaper, 10 pt, conference]{ieeeconf}
\IEEEoverridecommandlockouts 

\usepackage{graphics} 
\usepackage{float}
\graphicspath{ {images/} }
\usepackage{epsfig} 
\usepackage{times} 
\usepackage{amsmath} 
\usepackage{amssymb}  
\usepackage[separate-uncertainty=true]{siunitx}
\usepackage{physics}
\usepackage{mathrsfs}
\usepackage{hyperref}
\usepackage{caption}
\usepackage{rotating}
\usepackage[table]{xcolor}
\usepackage{bm}

\usepackage[export]{adjustbox}
\usepackage{subcaption}

\newcommand{\expect}[1]{\operatorname{\mathbb{E}} \left[ #1 \right]}

\title{\LARGE \bf
Monitoring the Mental State of Cooperativeness for Guiding an Elderly Person in Sit-to-Stand Assistance
}

\author{John Bell and H. Harry Asada
\thanks{John Bell and H. Harry Asada are with the Department of Mechanical Engineering,
        Massachusetts Institute of Technology, 77 Massachusetts Ave., Cambridge MA 02139, USA
        {\tt\small jhbell@mit.edu}
        {\tt\small asada@mit.edu}}%
}

\begin{document}
\bstctlcite{IEEEexample:BSTcontrol}

\maketitle
\thispagestyle{empty}
\pagestyle{empty}

\begin{abstract}
    In providing physical assistance to elderly people, ensuring cooperative behavior from the elderly persons is a critical requirement.  In sit-to-stand assistance, for example, an older adult must lean forward, so that the body mass can shift towards the feet before a caregiver starts lifting the body. An experienced caregiver guides the older adult through verbal communications and physical interactions, so that the older adult may be cooperative throughout the process. This guidance is of paramount importance and is a major challenge in introducing a robotic aid to the eldercare environment.
        
    The wide-scope goal of the current work is to develop an intelligent eldercare robot that can a) monitor the mental state of an older adult, and b) guide the older adult through an assisting procedure so that he/she can be cooperative in being assisted. The current work presents a basic modeling framework for describing a human's physical behaviors reflecting an internal mental state, and an algorithm for estimating the mental state through interactive observations. The sit-to-stand assistance problem is considered for the initial study.  A simple Kalman Filter is constructed for estimating the level of cooperativeness in response to applied cues, with a thresholding scheme being used to make judgments on the cooperativeness state.
\end{abstract}

\section{Introduction}
According to a report \cite{yi_characterization_2020}, over 80\% of those who died from COVID-19 in the United States were age 65 or older, and over 40\% of those individuals were either nursing home residents or staff. Additionally, as of April 28, 2021, it is estimated that over one million people in the US have been infected at nursing homes and long-term care facilities  \cite{times_nearly_2020}.  High COVID-19 transmission rates at nursing homes can likely be attributed to cluster infections caused by current in-person eldercare practices. The overarching goal of the current work is to develop a robotic aid for assisting older adults with virtually no direct physical interactions between an older adult and a caregiver.  

Existing elderly support technology has served primarily for reducing caregiver physical effort and making maneuvers more efficient.  One common example are ceiling lifts and floor lifters, such as the Arjo Maxi Sky \cite{arjo_nodate}, which are used to fully support the weight of elderly patients who cannot reliably walk, and transport them between seated positions at different locations.  Use of these devices still, however, requires in-person care; the caregiver has to lift the patient into and out of the lift, and to operate the lift's movement.  More recently, robotic devices have been developed to provide support to lift and lower a patient in/out of a seated position, such as the Fuji Hug \cite{fuji_nodate}.  To use such a device, the caretaker assists the patient in getting in the robot, which lifts the patient into an immobilized position; then, the caretaker moves robot and patient together.  While this does significantly reduce caretaker physical effort, it still requires sustained in-person interaction between patient and caregiver, which can be dangerous during epidemics.


We seek to enable methods of remote care in which a caretaker need not physically contact an elderly patient to assist with everyday actions.  
An elderly support robot would allow a caregiver or medical professional to remotely monitor the patient and remotely control the robot. No direct physical interactions between caregiver and patient are involved.  However, the cognitive workload could increase in operating the machine remotely, while physical workload reduces.  Furthermore, it is a challenge to provide the same level of quality care services from a remote site.  A fundamental issue in introducing a robotic aid is whether older adults accept to receive physical aids from robotic systems. Older adults may fear robotic aids. Physical aids cannot be provided safely and effectively, unless the older adult accepts the aid and behaves cooperatively. Care must be taken before executing a physical aid; the robotic system must make sure whether the older adult is ready to work with the robotic system; in other words, whether the older adult is cooperative.

This opens up a new research theme on human-robot interactions. Unlike the traditional framing of human-robot interaction problems \cite{blakemore_perception_2001,canessa_neural_2012,yu_human_2015,montesano_learning_2008}, the behavior of a human can be completely different depending on the conditions of the human.  An older adult may be cooperative or non-cooperative depending on his/her perception, mood, and other factors, which we describe as mental state. In the traditional setting, there is a premise or a consensus between a robot and the human, sharing a mission, task goals, and/or desired states. The challenge of eldercare is that it is uncertain how the human behaves; whether the human agrees and accepts the robotic aid, and cooperates with the robotic system. The robotic system must estimate and ensure the internal mental state of the human. 

In this context, two major challenges must be addressed in eldercare robotics. One is to estimate the mental state of an older adult through communication and interactions prior to execution of a physical aid. The other is to guide an older adult to be cooperative, persuading, convincing, or changing his/her mental state to agree, accept, and cooperate with the physical aid. Verbal communication, hand gestures, physical demonstration, and other means of cues must be given to older adults in order to guide their mental state. In an attempt to establish a new methodology, the current work presents a simple modeling and estimation framework in the specific context of sit-to-stand assistance. In the following, we will discuss how an older adult's behavior may differ depending on his/her mental state, how the characteristic behavior difference can be detected and, through verbal and light physical cues, how the mental state can be guided and confirmed. Based on these arguments, a simple model and estimation method will be presented. Initial human subject tests are conducted, and the proposed method is evaluated.

\section{Caretaker–Care Recipient Interactions in Sit-to-Stand Assistance}

\begin{figure}[h]
    \centering
    \includegraphics[width=3in]{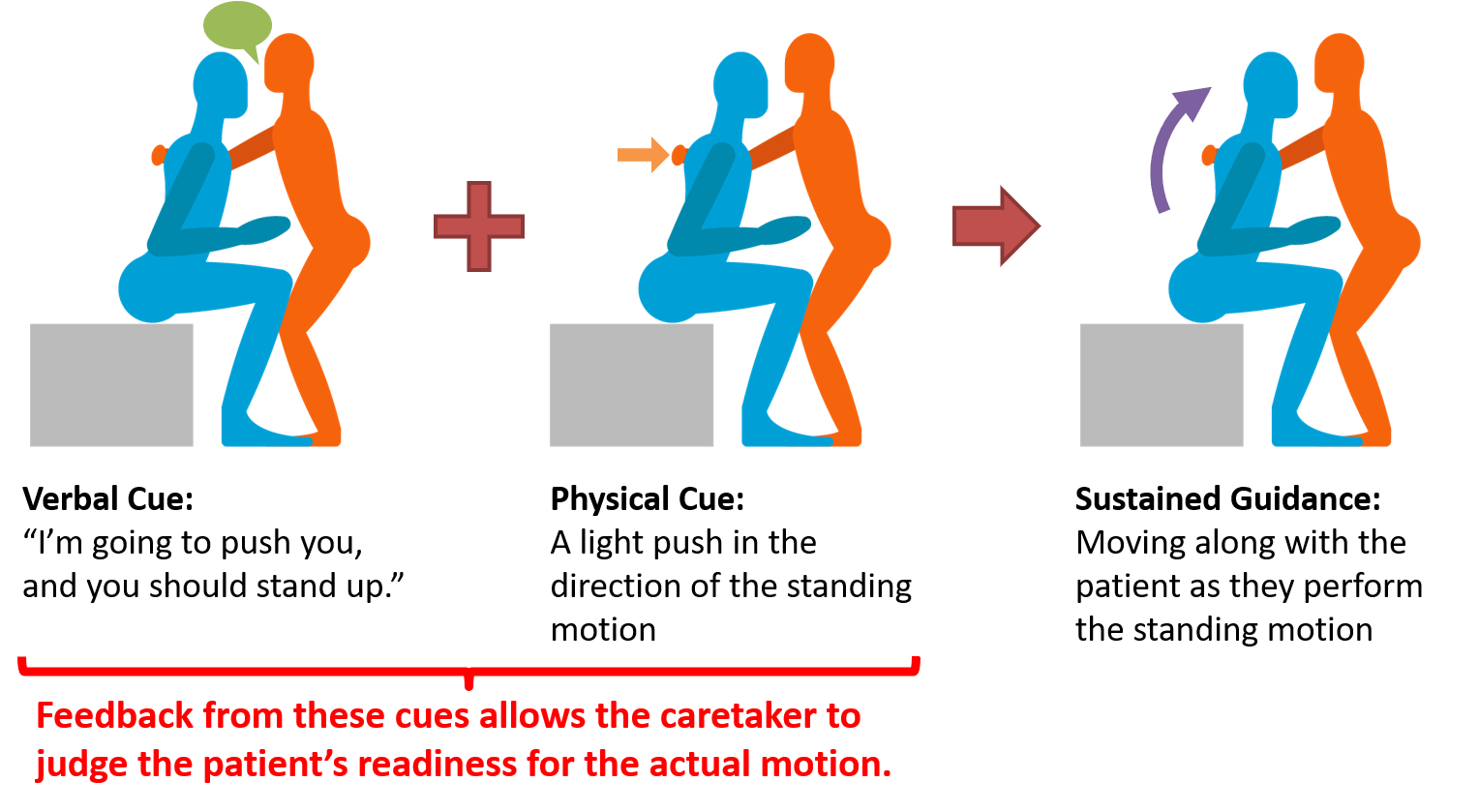}
    \caption{Illustration of the physical and verbal cues which are used by human caregivers to prepare a patient for a motion}
    \label{fig:human_cues}
\end{figure}

This section discusses interactions between a caretaker and a care recipient (i.e., an older adult to receive physical assistance). Assistance in the sit-to-stand transition is considered as an exemplary case study (see Fig. \ref{fig:human_cues}).  A caretaker needs cooperation from the care recipient for this class of physical assistance. The caretaker cannot execute a task without gaining a cooperative attitude and behavior from the care recipient \cite{nicholl_elderly_2012}. It is rude and even dangerous, if the caretaker abruptly yanks the older adult on his/her feet.  During the sit-to-stand transition, the care recipient first leans forward and bends the torso, so that the Center of Mass (CoM) of the upper body goes over the Base of Support (BOS) of the feet \cite{roebroeck_biomechanics_1994}. This allows the care recipient to maintain balance when leaving the seat. Instead, if the caretaker pulls the older adult although the CoM is away from the BOS, the older adult may feel that the caretaker is acting brutishly and dangerously, and may resist being assisted. In the case of a support robot, the older adult may be scared if the robot's action is unexpected and he/she is not ready to receive an assistive action.  These caretaker-care recipient interactions must be analyzed and understood in order to develop a functional support robot.

We begin with investigating the current practice of in-person assistance, and then derive a model describing the caretaker -- care recipient interactions. Eldercare handbooks and caretaker training manuals are useful resources for understanding the procedure and skills required for assisting older adults\cite{alwan_eldercare_2008,nicholl_elderly_2012,wick_wheelchair-bound_2007}. Based on years of experience in assisting various older adults with diverse physical and cognitive abilities, key techniques and specific procedures have been documented in these references [3]. A few important points we can learn from these are:
\begin{itemize}
    \item In providing older adults with physical assistance, it is critically important to explain the goal and procedure of assisting action; what the caretaker will be doing and what the older adult should expect. Before taking a physical action, the caretaker must let the older adult prepare for the action.
    \item It is not likely that an older adult voluntarily takes a cooperative action, e.g. bending forward, following the caretaker’s verbal explanation. Rather, older adults begin to take cooperative actions in response to the caretaker’s physical cues, e.g. gently pushing the back of the older adult. Physically touching, holding, and/or pushing the body of care recipient must be combined with verbal communication.
    \item The caretaker must assure that the care recipient is engaged and cooperative before taking the action. The caretaker must guide the care recipient with verbal communication and physical cues, so that he/she can be cooperative and ready for receiving assisting actions. 
\end{itemize}

The most challenging is to confirm whether the care recipient is cooperative and ready. This is to estimate the internal mind or mental state, indicating to which degree the care recipient is engaged, cooperative, and accepting being assisted. This mental state cannot be measured directly. However, in the context of sit-to-stand transition, it is conceivable that the care recipient's behaviors and responses to verbal and physical cues reflect the mental state of the care recipient. For example, as a caretaker guides the care recipient and gently pushes their back in order to prompt the care recipient to bend forward, the care recipient may comply to the gentle push if their own mental state is engaged and cooperative. If not cooperative and not feeling comfortable with the physical assistance, he/she may resist against being pushed and may even try to bend backward. Such physical responses to physical cues may reveal the mental state of the care recipient. In the current work, we hypothesize that cue--response behaviors reflect the mental state of a care recipient.

\begin{figure}[h]
    \centering
    \includegraphics[width=3in]{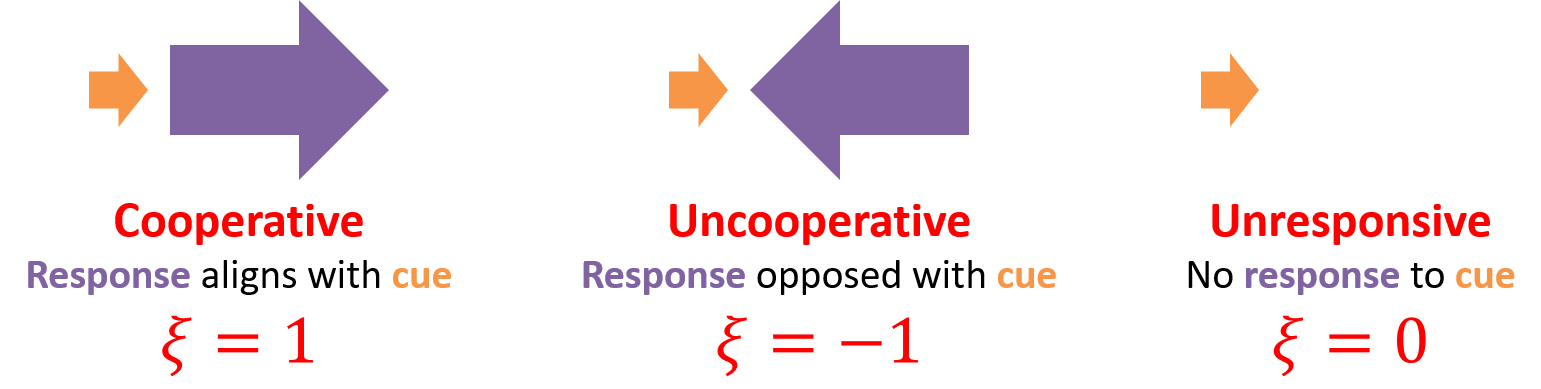}
    \caption{Illustration of the three states of cooperativeness, as they connect cues to their corresponding responses.}
    \label{fig:coop_summary}
\end{figure}

\section{Modeling}
Based on the hypothesis on cue--response behaviors reflecting a mental state, we aim to translate the current practice and skills of sit-to-stand assistance into a computable model to be identified from experimental data. The model will be used for guiding and estimating the cooperativeness mental state of a care recipient.

First, we represent the level of cooperativeness with a scalar variable $\xi$:

\begin{equation*}
     \xi = \begin{cases}
        +1 & \text{: Completely Cooperative} \\
        0  & \text{: Neutral / Unresponsive} \\
        -1 & \text{: Completely Uncooperative}
    \end{cases} 
\end{equation*}

This internal mental state influences observable behaviors in response to a caretaker’s cues. For the sit-to-stand assistance, we consider a gentle push of the care recipient’s back as a physical cue given by a caretaker. Thus, the physical cue $c_p$ has a unit of force.

In an attempt to connect the internal mental state $\xi$ to observable behaviors in response to physical and verbal cues, we elaborate a model representing the relationship based on the following considerations:


\underline{Model Considerations}
\begin{enumerate}
    \item The human acts voluntarily to move their actual position $x$ to their intended position $\eta$.
    \item Cooperativeness $\xi$ affects how the cues $c_p$ and $c_v$ lead to changes in the human's intended position $\eta$.  In the cooperative case ($\xi=1$), the change of $\eta$ induced by a cue is in the same direction as the cue.  The opposite is true for the uncooperative case ($\xi=-1$). A cue and the resultant change in $\eta$ are in the opposite directions. In the neutral, or unresponsive case ($\xi=0$), cues do not affect $\eta$.
\end{enumerate}

For Consideration 1, we can apply a neuromotor control and biomechanic model. Considering only 1 dimensional motion of the torso bending,
\begin{equation}\label{eq:base_dynamics}
    \ddot{x} = - \lambda_1 \dot{x} + k_1 \left( \eta - x \right) + k_2 \, c_p + w_{x,t}
\end{equation}
where $x$ is displacement of the torso, $c_p$ is the force applied by a caretaker, and $w_x$ is process noise. The physical sense of the three major terms is:
\begin{itemize}
    \item $-\lambda_1 \dot{x}_t$ - The effect of damping.  Parameter $\lambda_1$ is the damping rate of physical motion.
    \item $k_1 \left( \eta_t - x_t \right)$ - The voluntary acceleration of the care recipient, with the purpose of making $x \to \eta$.  This is of the form suggested by the equilibrium-point hypothesis of human motor control \cite{equilibrium-point}, and is equivalent to proportional control.  Parameter $k_1$ is the proportionality constant.
    \item $k_2 \, c_p$ - The involuntary acceleration due to being pushed by a physical cue.  Parameter $k_2$ is effective stiffness constant for this interaction.
\end{itemize}

The above equation is in the physical domain, hence no verbal cue is involved. The physical cue, although small in magnitude, is involved in pushing the system. Parameters $-\lambda$, $k_1$, and $k_2$ represent the impedance of the motor control system. In summary, the parameters for this physical model are collectively represented as:
\begin{equation}
    \theta_{dyn} = \left\{ \lambda_1, k_1, k_2 \right\}
\end{equation}

For Consideration 2, the intended position is generated in response to both verbal and physical cues. Unlike physical cues, verbal cues cannot be represented as continuous physical values. Rather, they are symbols and their effect upon changes to intended position $\eta$ is better represented in discrete time.

\begin{equation}
    \eta_{t+1} = f_{int}\left( \eta_t, c_{p,t}, c_{v,t}, \xi_t; \theta_{int} \right) + w_{\eta,t} 
\end{equation}
where subscript $t$ represents discrete time step, $f_{int}$ is a function with parameters $\theta_{int}$, and $w_{\eta,t}$ is noise.

This transition of the intended position significantly differs depending on the mental state.
As an example embodiment of this relationship, we can consider the following simple model where the variable of the mental state is multiplied to the combined term of the verbal and physical cues.
\begin{equation}\label{eq:intention_model_simple}
    \eta_{t+1} = \eta_t + \xi_t \cdot \left( k_3 \, c_{p,t} + k_4 \, c_{v,t} \right) dt + w_{\eta,t}
\end{equation}
There are two similar non-noise components to the $\eta$ update:
\begin{itemize}
    \item $k_3 \, \xi_t \, c_{p,t}$ - The effect of a physical cue $c_p$ on the change of intended position $\eta$.  Parameter $k_3$ is the effective gain of physical cues here.
    \item $k_v \, \xi_t \, c_{v,t}$ - The effect of a verbal cue $c_v$ on the change of intended position $\eta$.  Parameter $k_4$ is the effective gain of verbal cues here.
\end{itemize}
For both of these, cooperativeness affects this cue-resultant change of $\eta$ in the following ways:
\begin{itemize}
    \item $\xi = 1$ / Cooperative: Cues to move in a given direction change the human's intended position in the same direction.
    \item $\xi = -1$ / Uncooperative: Cues to move in a given direction change the human's intended position in the opposing direction.
    \item $\xi = 0$ / Unresponsive: Cues to move in a given direction do not change the human's intended position.
\end{itemize}
These behaviors are illustrated in Figure \ref{fig:coop_summary}. In summary, the parameters for this model stage are the following:
\begin{equation}
    \theta_{int} = \left\{ k_3, k_4 \right\}
\end{equation}



Finally, another model is necessary to represent how the mental state of cooperativeness is changed or guided towards the cooperative state by applying both physical and verbal cues.
\begin{equation}
    \xi_{t+1} = f_{coop}\left( \xi_t, c_{p,t}, c_{v,t}; \theta_{coop} \right) + w_{c,t}
\end{equation}
where $f_{coop}$ is a function of the current mental state $\eta_t$, and physical and verbal cues, and contains parameters $\theta_{coop} $, and $w_{c,t}$ is noise.


The proposed model serves to elucidate how the observable behaviors of a care recipient reflect the internal mental state of cooperativeness at three levels of physical and mental dynamics. If the sole objective is to let a care recipient take a bending posture, there is no need to estimate the internal mental state. Simply measuring the bending angle, the caretaker could execute an assisting procedure. However, the caretaker – care recipient relationship is more complex. The caretaker would make a costly mistake if he/she misunderstands that the care recipient is ready and cooperative. Two failure scenarios for such a na\"ive method are:

\underline{Potential Failure Scenarios}
\begin{itemize}
    \item As the caretaker applies a force, i.e. pushing the back of the care recipient, he/she may bend forward if the force is significantly large and/or the care recipient is too infirm to resist against the force. Although the care recipient is not ready or not accepting the physical assistance, he/she may bend forward.
    \item Due to uncertainties at all three levels, the care recipient may exhibit a cooperative behavior although he/she has not yet been ready.
\end{itemize}
Simply detecting a body movement alone may not be a reliable method for determining the cooperativeness state. The method should be robust against uncertainties. The judgement must be made not merely based on a snapshot observation; a series of observations and previous history of individual care recipients must be incorporated and exploited for reliable judgement.  The proposed model provides us with a framework for constructing a robust estimator.

\section{Cooperativeness Estimation Using a Kalman Filter with Thresholding}

A simplified linear model is used for constructing a Kalman Filter for estimating the cooperative mental state. Namely, we use \eqref{eq:base_dynamics} for the biomechanic model and \eqref{eq:intention_model_simple} for the intention model. As for the mental state transition, we use the following random process: 
\begin{equation}\label{eq:coop_random_process}
    \xi_{t+1} = \xi_t + w_{\xi,t}
\end{equation}
Due to only focusing on cooperativeness estimation, not cooperativeness prediction, for this portion of the analysis, a na\"ive model is used for the cooperativeness model stage.  Ideally, a rich predictive model for cooperativeness could be used to bolster the effectiveness of cooperativeness estimation; however, given the current lack of such a predictive model, the na\"ive random process of \eqref{eq:coop_random_process} can serve as a placeholder.


In order to run the Kalman filter, we can assemble the following linear time-variant state-space model to reflect our simplified version of the three-stage model:\\
\underline{State Update Model}
\begin{equation}
    \bm{x}_{t+1} = A\left(c_{p,t},c_{v,t}\right) \, \bm{x}_t + B \, c_{p,t} + w_t \text{  (noise)}
\end{equation}
where the state vector is $\bm{x}_t := \begin{bmatrix}x_t & \dot{x}_t & \eta_t & \xi_t \end{bmatrix}^T$, and the time-dependent parameter matrices are constructed from the three governing equations.\\
\underline{Measurement Model}
\begin{equation}
    y_t = H \bm{x}_t + v_t \text{  (noise)}
\end{equation}
\begin{equation}
    H = \begin{bmatrix}1 & 0 & 0 & 0\end{bmatrix}
\end{equation}

\subsubsection{Cooperativeness Judgment with Thresholding}
\begin{figure}[h]
    \centering
    \includegraphics[width=2.5in]{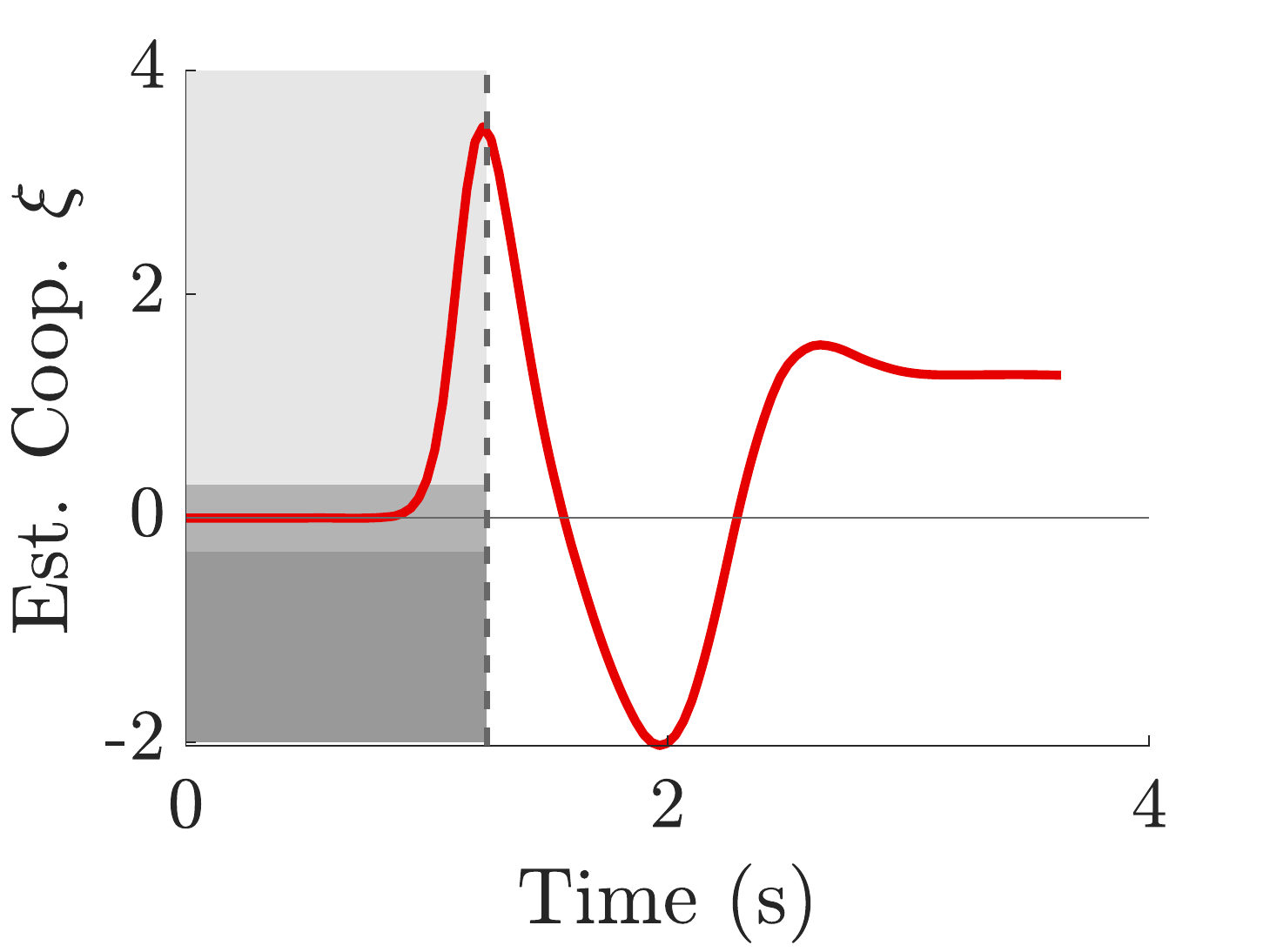}
    \caption{Demonstration of the judgment threshold.  Here it is performed 0.7 seconds after the initial cue (1.2 seconds on the time axis), with a threshold of $\xi = \pm 0.3$.  The cooperativeness is judged to be $\xi = 1$ (lightest gray region).}
    \label{fig:xi_threshold}
\end{figure}

The Kalman filter is a continuous-domain filter; as such, it generates a continuous estimate $\hat{\xi}$.  Nevertheless, cooperativeness $\xi$, as defined, is discrete, with a domain of $\xi \in \{-1,0,1\}$.  Practical judgments of cooperativeness $\xi$ should be made discretely.  To bridge this gap, we propose the following thresholding method: at a given time after the initial onset of an applied cue, the Kalman-filter estimated value of $\xi$ is sampled, then compared to a threshold $\xi_{thresh}$.  If $\hat{\xi}\geq\xi_{thresh}$, then cooperativeness is judged to be $\xi = 1$.  If $\hat{\xi}\leq-\xi_{thresh}$, then cooperativeness is judged to be $\xi = -1$.  If $-\xi_{thresh}<\hat{\xi}<\xi_{thresh}$, then cooperativeness is judged to be $\xi = 0$.  This scheme is illustrated in Fig. \ref{fig:xi_threshold}.

\section{Experimentation}
We sought to verify the efficacy of this Kalman-Filter-based cooperativeness estimation technique, using basic experimental testing.  In this stage of experimental testing, we focused on characterizing response to visual and physical cues given by a human caregiver to a human test subject.  This test is meant to be the basis for development of sensing methods that could be used by a robotic elderly assist device; as such, though the cues were given by a human, we limited the observables to those that could be measured by a robotic system.  As the main purpose of these experiments is to reinforce the mathematical basis for a cooperativeness estimation algorithm, not to establish the actual behavioral patterns of potential patients, we deemed it unnecessary to request the participation of elderly or physically disabled test subjects.  We ran the following test on four healthy young adult human subjects:

\subsection{Experimental Procedure}

\begin{figure}[h]
    \centering
    \vspace{0.2in}
    \includegraphics[height=1.5in]{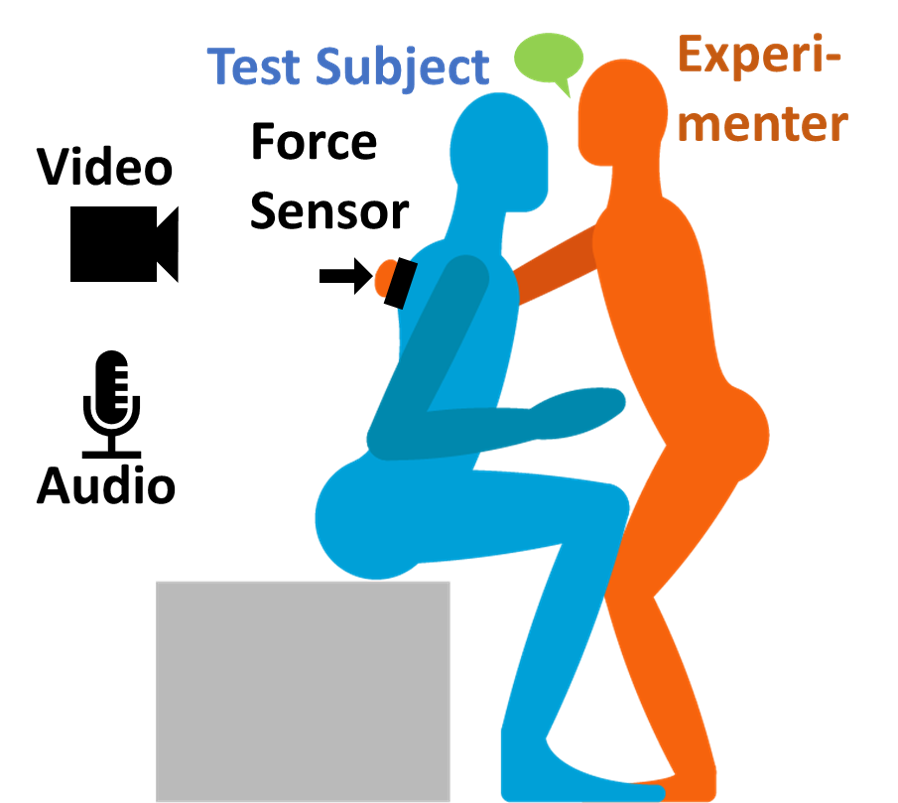}\includegraphics[height=1.5in]{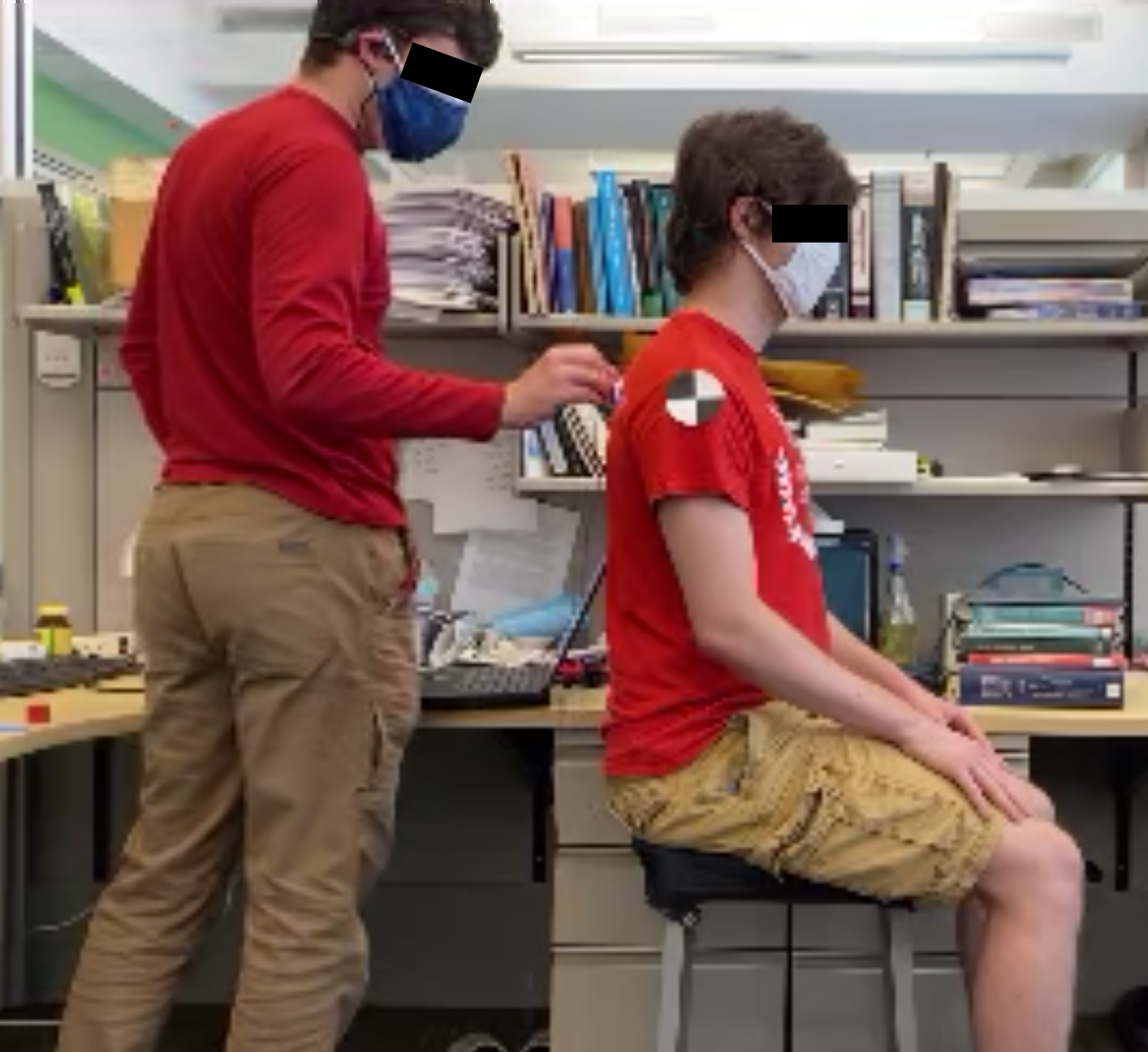}
    \caption{Left: Illustration of the first stage of experimental setup used to collect human cue response data. Right: Video frame of the experimental procedure, taken during the execution of an experimental trial.}
    \label{fig:experimental_setup}
\end{figure}
\begin{figure}[h]
    \centering
    \includegraphics[height=2em]{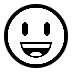}\quad\includegraphics[height=2em]{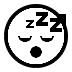}\quad\includegraphics[height=2em]{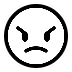}
    \caption{The emojis used to communicate to the test subjects the approximate meanings of the three cooperativeness states.  Left to right: Cooperative, Unresponsive, Uncooperative}
    \label{fig:emojis}
\end{figure}
\textit{This procedure was determined to be exempt by MIT's IRB, COUHES, on grounds of being a benign behavioral intervention (MIT COUHES Exempt ID E-3420).}

\textit{For reference, this experimental setup is illustrated and photographed in Fig. \ref{fig:experimental_setup}.}

Before the test, the human test subject is explained the general purpose of the experiment, but not the details of the cue response model.  The human test subject is seated on a stool.  To request a specified cooperativeness state of a given trial, the experimenter provides the human test subject with a card which displays the word ``Cooperative", ``Unresponsive", or ``Uncooperative", as well as an emoji which approximately represents the emotional context of the requested cooperativeness state (see Fig. \ref{fig:emojis}).  The subject is NOT explained the modeled definitions of the cooperativeness states, in the concern that awareness of the model may influence their behavior.

The experimenter then proceeds to provide the test subject with a physical cue: a push forward to the upper back, a verbal cue: a command to move forward or backward, or both.  The human test subject responds as they see fit, given the requested cooperativeness state provided by the experimenter beforehand.  The physical cue is delivered though an Optoforce compressive force sensor held by the experimenter; thus, the force magnitude of the physical cue is measured.  The timing and direction of the verbal cue is recorded using a microphone.  The response of the human test subject is characterized in terms of the horizontal position of their shoulder; this is measured using a video camera, which tracks a visual target taped to the subject's shoulder.

\subsection{Experimental Parameters}
Four subjects were tested, in a series of 25 different trials.  Each of the 25 trials was randomly assigned one of each of the following properties:
\begin{itemize}
    \item Requested Cooperativeness: $\xi\in\{-1,0,1\}$
    \item Physical Cue Intensity: \{None, Soft, Hard\}
    \item Verbal Cue: \{None, Move Back, Move Forward\}
    \item Relative Timing Between Physical and Verbal Cues (if both): $\{-4,-3,-2,-1,0,1,2,3,4\}$ seconds
\end{itemize}
The same set of trial parameters was used for each human test subject.

\subsection{Experimental Results}
\subsubsection{Kalman Filter Performance Analysis}
\begin{figure}[h]
    \centering
    \vspace{0.2in}
    \includegraphics[width=1.7in]{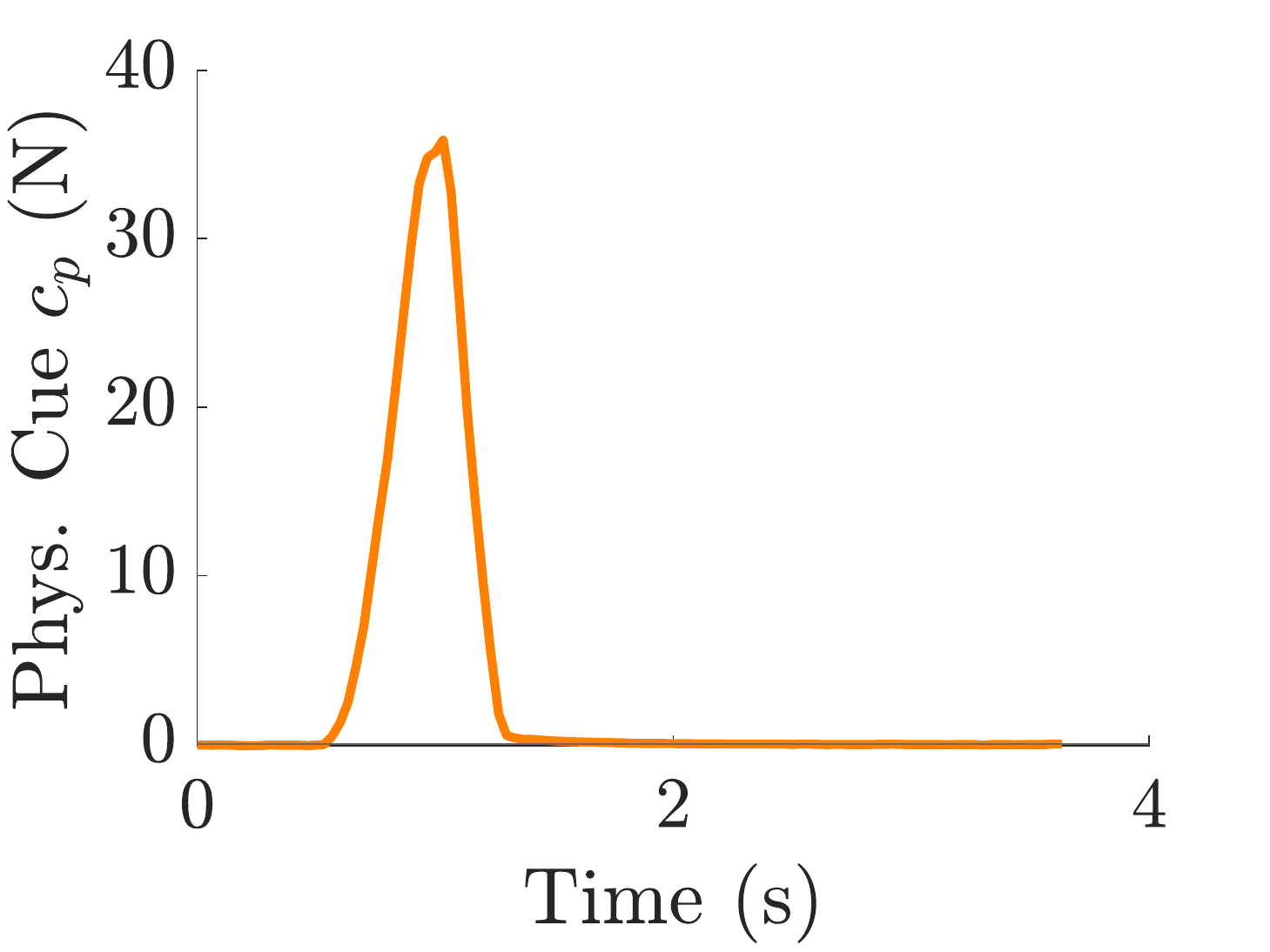}\includegraphics[width=1.7in]{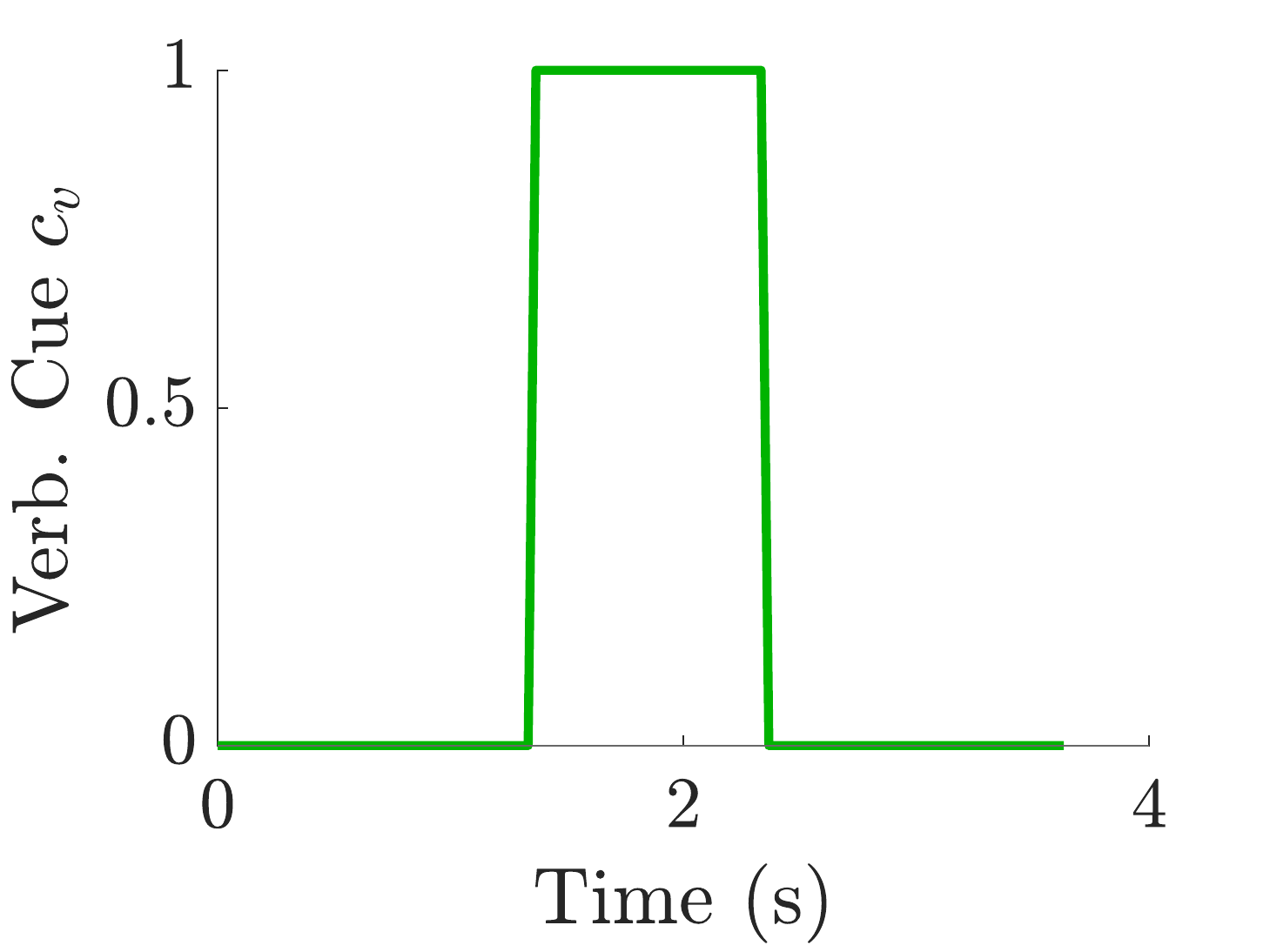}\\
    \includegraphics[width=1.7in]{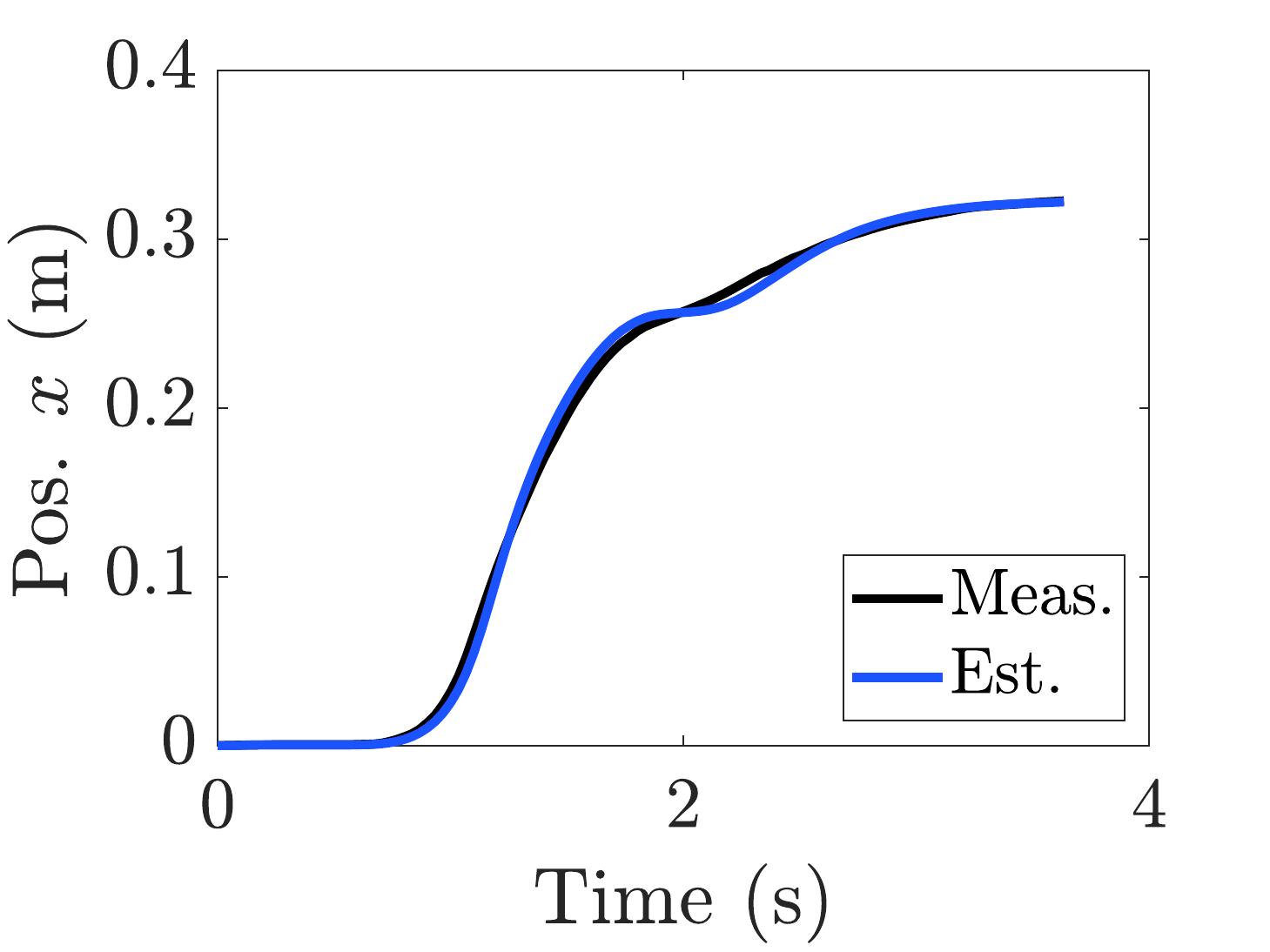}\includegraphics[width=1.7in]{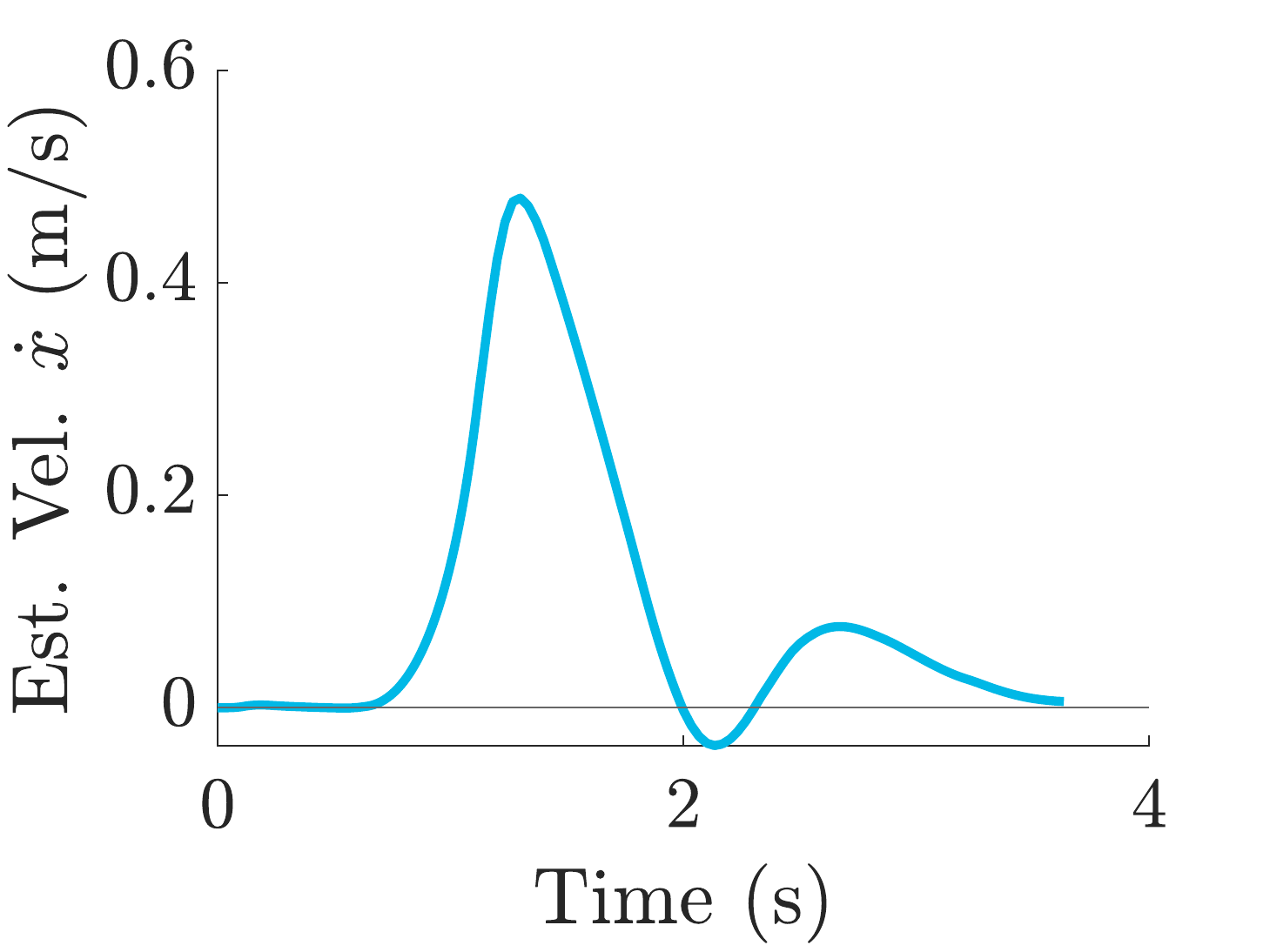}\\
    \includegraphics[width=1.7in]{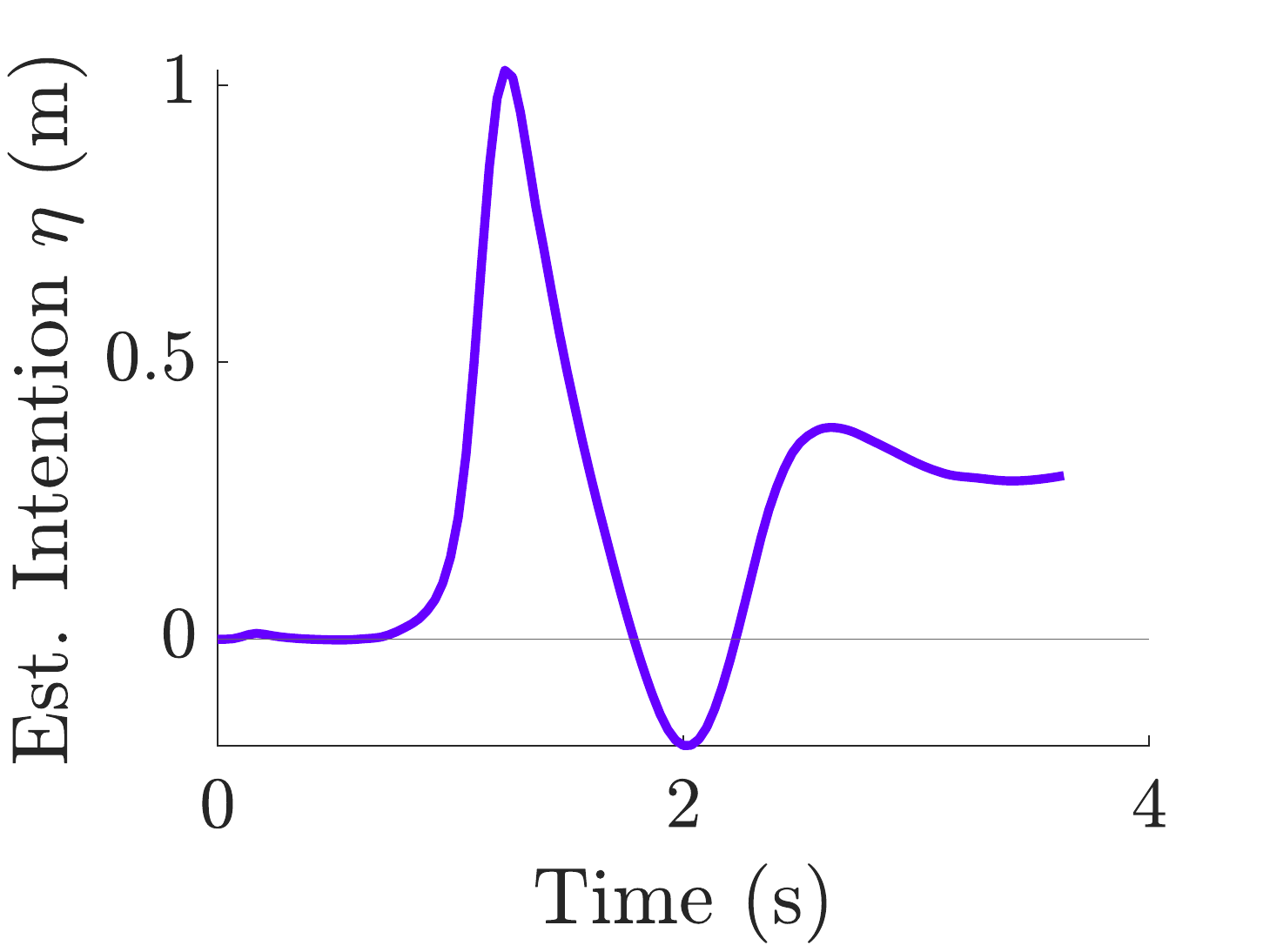}\includegraphics[width=1.7in]{1_3_xi.pdf}
    \caption{Response of Subject 1 to both a forward push cue and a ``Move Forward" verbal cue, with a requested cooperativeness of $\xi = 1$.  Given the measured cues $c_p$ and $c_v$, and measured position $x$, the human subject's modeled position $x$, velocity $\dot{x}$, intended position $\eta$, and cooperativeness $\xi$ are estimated using a modified Kalman filter.  The thresholds and time window for cooperativeness judgment are shown in the plot for estimated $\xi$.}
    \label{fig:data_kalman}
\end{figure}

The results of a single trial are shown in Figure \ref{fig:data_kalman}.  In this trial, Subject 1 was requested to be cooperative ($\xi=1$).  Subsequently, a hard forward physical cue was applied, immediately followed by a verbal cue to move forward.  Subject 1 was accelerated directly by the push, and then proceeded to move themself forward further.  As shown in Fig. \ref{fig:data_kalman}, the Kalman filter estimates two pulses in intended position $\eta$, an initial large, brief pulse corresponding to the initial acceleration, and a second smaller, sustained pulse corresponding to the additional motion.  This two-pulse estimation of intended position $\eta$ leads to a similar two-pulse estimation of cooperativeness $\xi$, which briefly goes negative between the pulses.

This brief negative estimate of $\xi$ (which should be $\xi = 1$ throughout) is due to one of the main disadvantages of this Kalman filter: it cannot perfectly distinguish the direct dynamic motion caused by hard physical cues from the indirect voluntary motion caused through cooperativeness.  While $k_2 \, c_p$, the direct mechanical compliance term of the dynamics equation \eqref{eq:base_dynamics}, does serve to predict the $x$ motion directly caused by a physical push, any variation in the effective mechanical compliance of the human subject will result in $x$ motion being attributed to intended motion $\eta$ instead.  This can be seen in Fig. \ref{fig:data_kalman}, where the initial acceleration is attributed to an unusually large spike in intended position $\eta$ of 1 meter.  Nevertheless, the fact that Subject 1 does not resist the motion results in their cooperativeness being correctly estimated as $\xi = 1$ by the threshold judgment.  The threshold judgment for all data was made at a delay of 0.75 seconds after cue initiation, and a threshold of $\xi = \pm 0.3$.

\subsubsection{Aggregate Analysis of Cooperativeness Judgment}

\begin{table}[h]
    \centering
    \includegraphics[width=3in]{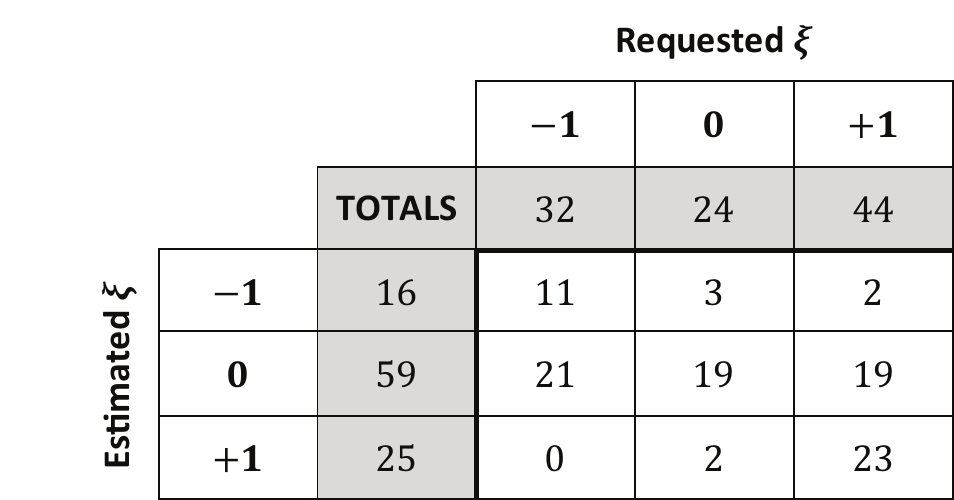}
    \caption{Number of trials across all test subjects ($N=100$), sorted by their requested (true) value of cooperativeness $\xi$ and their estimated cooperativeness $\xi$, as estimated by the Kalman filter with threshold judgment.}
    \label{tab:raw_counts}
\end{table}

While only one trial is presented in Fig. \ref{fig:data_kalman}, there were 100 different trials actually performed across the 4 human test subjects.  For the sake of brevity and ease of reading, aggregate data is presented only in Tables \ref{tab:raw_counts}--\ref{tab:obs_counts}.  Table \ref{tab:raw_counts} shows the overall results of the thresholded Kalman filter cooperativeness judgment scheme; specifically, it sorts the 100 trials in terms of the cooperativeness value requested of the test subject (the effective ground truth) and the cooperativeness value estimated by the judgment scheme.  This table effectively shows the joint sample distribution of requested $\xi$ and estimated $\xi$.

\begin{table}[h]
    \centering
    \vspace{0.1in}
    \includegraphics[width=2.5in]{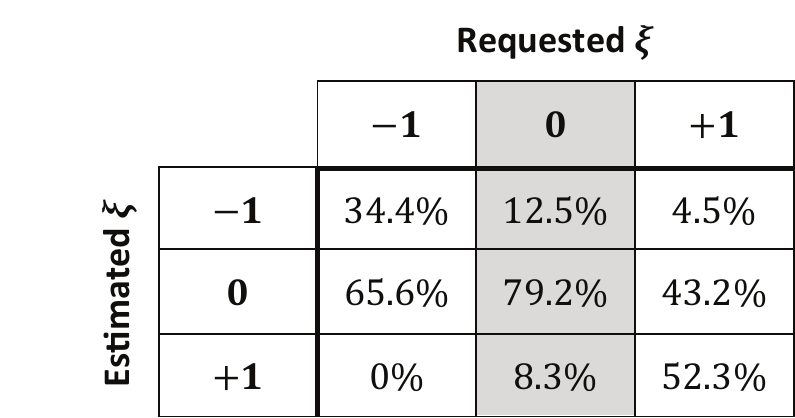}
    \caption{Conditional sample distribution of estimated cooperativeness $\xi$, conditioned on the requested (true) cooperativeness $\xi$}
    \label{tab:req_counts}
\end{table}

These results can also be analyzed in terms of conditional sample distributions.  Table \ref{tab:req_counts} shows the sample conditional distribution of estimated $\xi$ given requested $\xi$.  In other words: if a trial if known to have a certain requested $\xi$, what is the sample probability of it having been estimated to be a certain value?  The judgment scheme is very good at judging unresponsive cases to be unresponsive, at a rate of 79.2\%.  An uncooperative cases is most likely to be judged as unresponsive, with a dismal correct rate of judgment of 34.4\%.  Cooperative cases are most likely to be judged as such, with a correct rate of judgment of 52.3\%.

\begin{table}[h]
    \centering
    \includegraphics[width=2.5in]{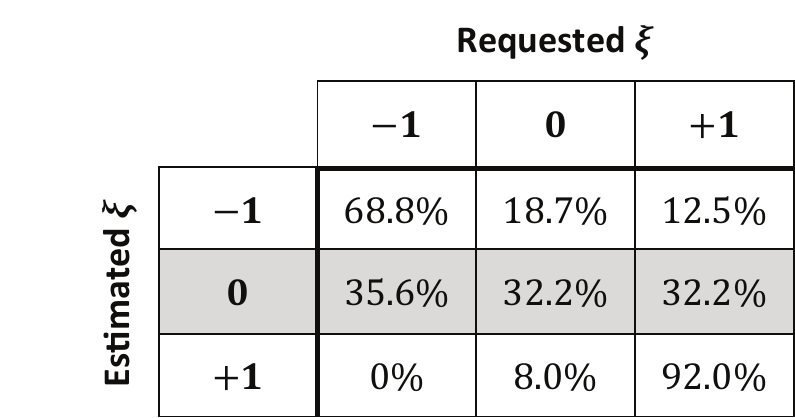}
    \caption{Conditional sample distribution of requested (true) cooperativeness $\xi$, conditioned on the estimated cooperativeness $\xi$}
    \label{tab:obs_counts}
\end{table}

The dual conditional analysis also sheds some valuable light.  Table \ref{tab:obs_counts} shows the sample conditional distribution of requested $\xi$ given estimated $\xi$.  In other words: if a trial is estimated to have a given $\xi$, what is the sample probability of it having actually been requested to be a given value $\xi$?  The judgment scheme is relatively trustworthy in judgments of being cooperative, with a $92\%$ true positive rate.  Judgments of uncooperative are more likely than not to actually be uncooperative, with a $68.8\%$ true positive rate.  Judgments of unresponsiveness mean effectively nothing, however, with a roughly equal chance of having been caused by any requested cooperativeness.

\section{Conclusion}
This paper has addressed one of the key challenges in applying robotic systems to the eldercare environment. To be viable, the technology must be accepted by elderly people, and robots must be able to gain a cooperative behavior from the elderly. We have presented a basic modeling framework for estimating the mental state of cooperativeness in the context of sit-to-stand assistance. From existing literature and practical eldercare knowledge, we have found that the internal mental state of an older adult is reflected to his/her observable behaviors in response to verbal and physical cues given by a caretaker. Based on this, a mental-physical model has been constructed at three levels. The lowest is a biomechanic, neuromotor control model relating observable behaviors to his/her intended posture and movement. The highest is a mental model indicating his/her cooperativeness, which can change depending on a caretaker's guidance, i.e. verbal and physical cues. The middle layer is a model linking the mental state to intended posture and movement. Based on this model, a thresholded Kalman filter judgment scheme has been constructed to estimate the mental state from observable body responses to a caretaker's cues. IRB-approved human subject tests were conducted to verify the efficacy of this judgment algorithm in healthy adults. The thresholded Kalman filter cooperativeness judgment scheme proposed in this work successfully identified cooperative response, with a 92\% true positive rate.





\bibliographystyle{IEEEtran}
\bibliography{IEEEabrv,myBib}

\end{document}